\documentclass[AMA,Times1COL]{WileyNJDv5}

\articletype{Article Type}%

\received{Date Month Year}
\revised{Date Month Year}
\accepted{Date Month Year}
\journal{Journal}
\volume{00}
\copyyear{2023}
\startpage{1}

\begin{document}

\title{Automated Multilabel Mpox Research Classification with Explainable Transformer Models }    

\author[1]{Tanjim Taharat Aurpa}

\address[1]{\orgdiv{Department of Data Science and Engineering}, \orgname{University of Frontier Technology, Bangladesh (UFTB)}}

\corres{Corresponding author Tanjim Taharat Aurpa, This is sample corresponding address. \email{aurpa0001@uftb.ac.bd}}


\abstract[Abstract]{
The Mpox outbreak remains a serious public health issue, with the WHO (World Health Organization) reporting increasing cases in some regions. Research on Mpox is vital for several reasons, including vaccine development, diagnostic improvement, viral evolution studies, and preventing future outbreaks. However, the large amount of research being published makes it difficult to organize and analyze information efficiently.
This study focuses on using multilabel classification to categorize 14590 Mpox research articles into key topics such as outbreaks, vaccination, and epidemiology. Among the different AI models tested, BERT performed the best, achieving 97.05\% accuracy, 97.67\% micro F1 score, and 96.46\% macro F1 score. To better understand how the model makes decisions, SHAP was used to analyze significant word features and patterns.
The results show that BERT can help automate the classification of Mpox research, making it easier for researchers, policymakers, and healthcare workers to quickly find relevant information, saving time and improving public health efforts.}


\maketitle

\section{Introduction}
MPox, also known as Monkey Pox, is a harmful disease caused by the monkeypox virus. A Mpox career can be responsible for transmitting the virus to someone else. According to WHO\footnote{https://www.who.int/news-room/fact-sheets/detail/mpox}, 1951 was the year when this virus was invented in Denmark for the first time. After that, many countries have seen a devastating outbreak of this disease. Recently, a variant of Mpox has been spread over 120 countries from May 2022 to August 2024. During this outbreak, more than one lac laboratory-confirmed case was found, and 220 people died from being infected with Mpox. Recent research has identified multiple variants of the Mpox virus, emphasizing the evolving nature of the disease and its implications for public health. 
Two primary clades, Clade I (Central African) and Clade II (West African), have been discovered. However, Recent subvariants Clade Ib and IIb have combined and show various transmission dynamics and virulence. The recent surge in cases, particularly in Africa, has been alarming, with Mpox being declared a public health emergency and the WHO showing international concern in 2024. The spread into regions with limited previous exposure, such as East African nations like Burundi, Rwanda, and Uganda, highlights the virus's adaptability. Therefore, these evolving variants underline the need for enhanced surveillance, vaccination strategies, and public health interventions to control the spread of Mpox and mitigate its impact.

Mpox research is fundamental because the virus can cause large outbreaks, especially in places where healthcare systems are underdeveloped. Learning about the virus and creating ways to prevent and treat it can significantly reduce its impact on global health. Moreover, this research can find a way to offer cheaper and more efficient vaccines and 
With more animal diseases jumping to humans, studying Mpox helps us prepare for future infections. Recent outbreaks in new regions have highlighted how urgent this research is to protect public health and stop pandemics before they start. Existing research focuses mainly on outbreaks of this virus, prevention, and vaccinations. Again, one paper can relate to more than one of these subtopics.
This work will benefit researchers by proposing better public health policies, improving how we respond to outbreaks, and making us more ready for viral epidemics by utilizing Natural Language processing.

Utilizing modern transformer-based architecture such as BERT RoBERTa, AlBERT, DistillBERT, ELECTRA, and XLNet to analyze recent research on MPox is the primary motivation of this paper. These architectures have provided remarkable performance in different NLP tasks \cite{sy2024beyond}, \cite{aurpa2024ensemble}, \cite{cheruku2024sentiment}, \cite{aurpa2024instructnet} etc. These transformer models work in two steps: model pretraining and fine-tuning with downstream tasks. These models utilized numerous multi-label classification-based types of research. 
Multi-label classification, an NLP problem where a given input text is classified as more than one class, is a versatile tool. Transformer-based architectures have shown significant performance in this area, particularly for fine-tuning downstream tasks. As mentioned earlier, three major research classes were identified, and are 'Prevention-related,' 'Outbreak-related,' and 'Vaccine-related.' A single research can be conducted on all of these topics, demonstrating the adaptability of multi-label classification to diverse research topics. This makes it a valuable and time-saving tool for healthcare professionals, researchers, and public health authorities.

Explainable AI(XAI) is a valuable technique that helps researchers disclose deep learning black boxes. One of the popular XAI techniques, SHapley Additive exPlanations(SHAP), helps to understand important features that contributed to the prediction, model's transparency, error calculation, etc. SHAP is widely utilized in various domains, including text classification \cite{tao2023making}, multimodal prediction \cite{el2021multilayer}, and image processing \cite{ukwuoma2025enhancing}, among others. In this work, SHAP will help to understand how and for which words the research articles are classified.

The main goal of this work is listed below:

\begin{itemize}
\item Utilization of modern NLP techniques by proposing a multi-label classification approach for classifying MPox research articles.
\item To explore the effectiveness of transformer-based models in classifying MPox research articles into outbreak, prevention, and vaccine categories.
\item Collecting research abstracts related to MPox from PubMed\footnote{https://pubmed.ncbi.nlm.nih.gov/} and analyzing them to bring out significant insight for researchers.
\item Understand the model's transparency by understanding the most important word features for each class by applying the XAI technique SHAP.
\end{itemize}

The research questions which this work will address are mentioned below:

\begin{itemize}
   \item[1]How effective are transformer-based architectures (e.g., BERT, RoBERTa) in performing multi-label classification for MPox-related research into categories such as outbreak, prevention, and vaccine? 
 \item[2]What insights can SHAP provide about the critical features (e.g., words or phrases) that drive the classification of MPox-related research articles? 
 \item[3] What are the challenges and solutions in applying multi-label classification to overlapping topics within MPox research?
 \item[4]How can multi-label classification and XAI techniques improve public health strategies and decision-making for managing MPox outbreaks? 
\end{itemize}

\section{Literature Review}

Because of BERT's exceptional success in addressing a variety of difficulties, it has drawn a lot of interest from researchers. For instance, \cite{utka2020pretraining} focused on a Twitter dataset, pretraining BERT, to improve Latvian sentiment analysis for tweets. Furthermore, BERT has shown its efficacy in fake news detection. In \cite{rai2022fake}, the researchers proposed a hybrid approach by integrating BERT with an LSTM layer, resulting in a 2.50\% accuracy increase on the PolitiFact dataset and a 1.10\% increase on the GossipCop dataset. Additionally, \cite{aurpa2024ensemble} explored an ensemble of two BERT models for recognizing medical entities, achieving a remarkable accuracy improvement of 11.80\%.

In \cite{aurpa2026transparent}, the authors employed SHAP after classifying the relations between two math entities, using various transformer-based architectures to highlight the key features that influence predictions. Rabbi et al. applied SHAP for multi-label text classification on COVID-19-related texts, where Random Forest and BERT emerged as the top-performing classifiers \cite{rabby2023multi}. Additionally, SHAP has been used in fake news detection, authentic news recognition, and question classification tasks, as explored in \cite{tao2023making}, covering binary, multi-class, and multi-label classification scenarios in order to demonstrate explainability.

MPox is one of the current significant concerns for WHO. Thus, besides other domains there, much work has been conducted—mostly sentiment analysis on social media text has been performed by authors.
In \cite{thakur2023sentiment}, authors used the hydrator to collect 61,862 tweets related to MPox and applied VADER for sentiment analysis. Their analysis shows that almost half of the observations (46.88\%) tweets are negative, and the remaining tweets are positive and neutral.
Another work \cite{bengesi2023machine} is where authors collected 50000 multilingual tweets related to MPox. Here, they used not only VADER but also TextBlob to determine sentiment labels, and then they applied different machine learning models. They achieved the highest 0.9348 accuracy with the TextBlob annotation and CountVectorizer as the vectorization technique. Some work has been performed on topic modeling of mpox-related text. Thakur et al. cite{thakur2023analyzing} performed topic modeling using 601,432 Tweets and determined a total of 50 topics. The authors analyzed the average coherence and found the highest confidence value for the topic 'Views and Perspectives about MPox.'
In \cite{edinger2023misinformation}, public health massaging and misinformation are analyzed using MPox-related tweets using S-BERT, PCA, and UMAP. 
A total of 125,424 MPox-related texts are collected, and Latent Dirichlet Allocation is used to determine significant topics. Using the CamemBERT, they analyzed the sentiment \cite{nia2023mpox}.
Hajjo et al. (2025) \cite{hajjo2025advancing} comprehensively review Mpox epidemiology, diagnostics, and treatments, mapping critical host-pathogen interactions to reveal how the virus evades innate immunity. Their network pharmacology analysis identifies key host protein targets for novel drug discovery while highlighting the role of AI and big data in monitoring viral mutations.
The study \cite{Papageorgiouanovel} proposes a novel epidemiologically informed particle filtering model (EI-PF) that integrates a deterministic SEIRD framework with stochastic parameter estimation and a unique bivariate Poisson distribution. By incorporating penalty factors to enforce epidemiological constraints, the model eliminates irregular particle trajectories and achieves superior fitting and predictive accuracy on the 2022 U.S. monkeypox outbreak data compared to standard alternatives.
Highlighting the global impacts of recent mpox outbreaks, Hayman et al. \cite{Hayman2025mpox} outline how environmental disruption, forest fragmentation, and regional socio-political factors escalate zoonotic spillover risks. They advocate for a transdisciplinary One Health framework that prioritizes sustainable ecological practices, local community trust, and global resource equity to mitigate future epidemics.
This paper \cite{Papageorgioudynamic} introduces a stochastic SPIR compartmental model based on a 3-dimensional Markov chain to derive algorithmic formulas for key epidemic descriptors, such as outbreak size, transmission source impact, and time until death. By integrating an augmented state-space formulation with particle filtering to estimate time-varying parameters from empirical mpox data, the authors significantly improve predictive accuracy and online outbreak assessment compared to traditional constant-parameter methods.

\section{Materials and Methods}
\subsection{Preliminary Concepts}
\subsubsection{\textbf{Bidirectional Encoder Representations from Transformers(BERT)}}
\begin{figure}
    \centering
    \includegraphics[width=0.75\linewidth]{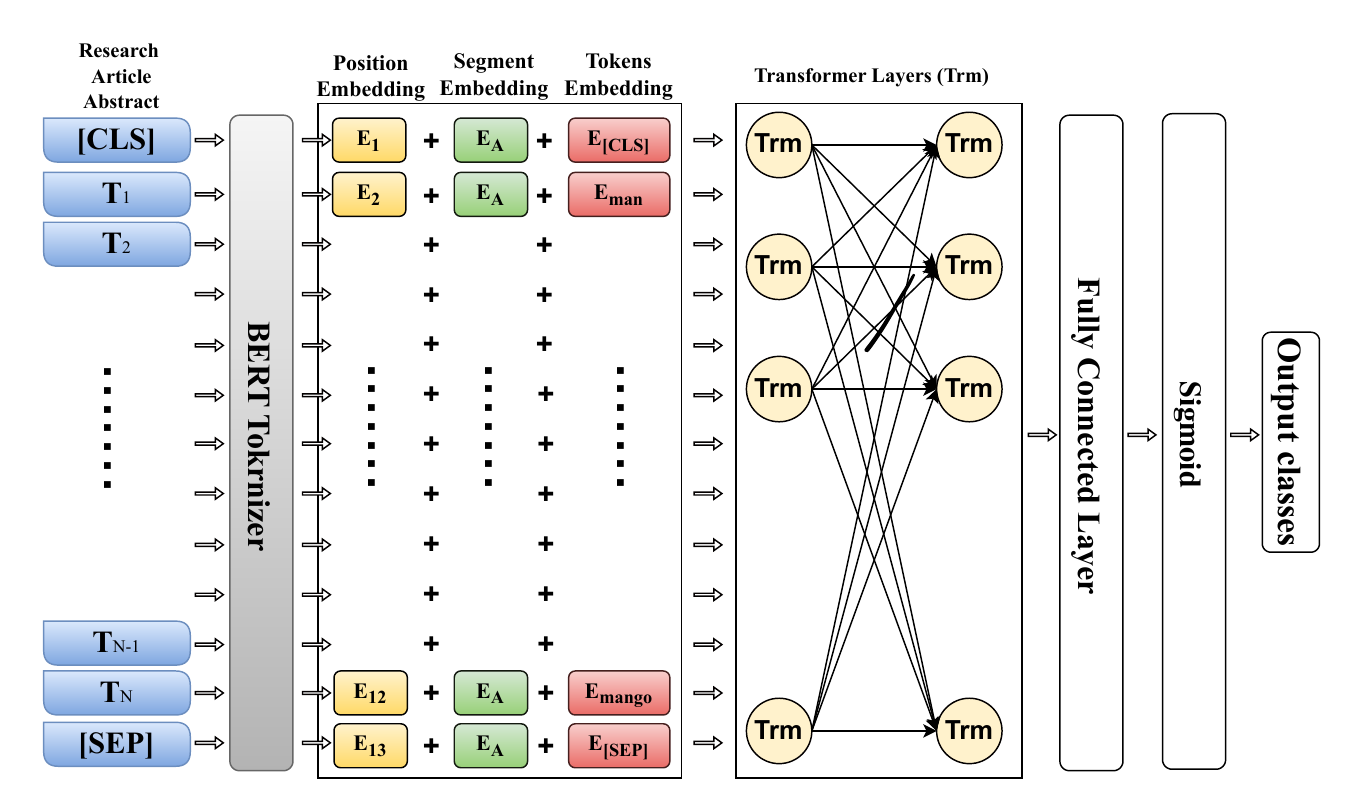}
    \caption{Proposed BERT Model}
    \label{fig:bert model}
\end{figure}

This study makes the utilization of Bidirectional Encoder Representations from Transformers (BERT). The attention mechanism is implemented in this transformer-based architecture. The two primary components of BERT are covered below:
\begin{itemize}
    \item \textbf{Pre-training BERT: } BERT is pre-trained using two primary tasks: Masked Language Modeling (MLM) and Next Sentence Prediction (NSP). In MLM, random tokens within the text are masked, and the model's goal is to predict these tokens, which helps it learn a bidirectional representation of language. NSP, on the other hand, enables the model to understand the relationships between sentences by determining whether one sentence logically follows another. BERT's training data consists of English Wikipedia (excluding lists, headings, and tables) as well as the BooksCorpus, which contains 800 million words.
    \item \textbf{Fine-tuning BERT: }BERT is often used for downstream tasks that involve either single sentences or pairs of sentences as inputs. It begins with pre-trained parameters that capture general language understanding and is subsequently fine-tuned on labeled data for specific tasks, such as classification, question answering, or sentiment analysis. In this work, BERT is fine-tuned for multilabel text classification, which predicts suitable labels for research papers.
\end{itemize}
Figure \ref{fig:bert model} depicts the architecture of the BERT model for multilable research article classification.

\subsubsection{\textbf{SHapley Additive exPlanations(SHAP)}}
SHAP(SHapley Additive exPlanations) is a powerful tool that helps us understand which features are most important in our model's decisions. It works by assigning an importance score to each feature based on Shapley values, a concept from game theory that ensures a fair distribution of credit among team members—except here, the "team members" are our features. In our case, SHAP helps explain how different aspects of research abstracts play a role in identifying and extracting key classes.

This paper incorporates SHAP to explain the decision-making process of the multi-label classification of research articles. By analyzing SHAP values, the most influential features in the model's predictions can be identified, along with their role in recognizing entities and their relationships. This approach enhances transparency in classification outcomes and helps refine the model by emphasizing the most important features.
\begin{itemize}
\item \textbf{Feature Importance:} SHAP helps identify which features, such as specific words, play the biggest role in accurately determining the classes.
\item \textbf{Model Transparency:} Using SHAP makes the model more understandable by revealing the most important features and the reasons behind its predictions. These explanations improve its reliability, making it a useful tool for improving decision-making.
\item \textbf{Error Analysis:} SHAP helps to indicate mistakes by showing which features cause incorrect predictions, allowing for targeted improvements to the model.
\begin{figure}
    \centering
    \includegraphics[width=0.5\linewidth]{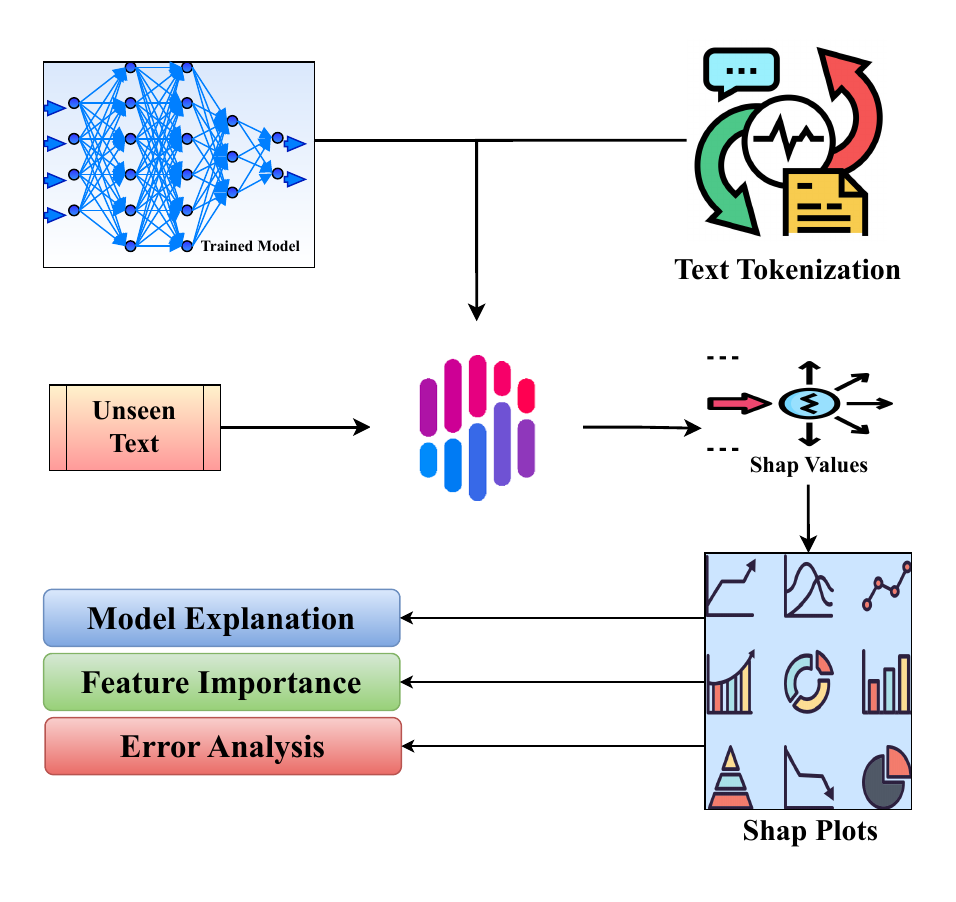}
    \caption{The working principle of SHAP algorithm}
    \label{explainability}
\end{figure}
Figure \ref{explainability} depicts the working process of the SHAP algorithm.

\end{itemize}
\subsection{Experimental Setup}
\subsubsection{Experimental Environment}
Deep learning models generally require high-performance computational resources to support parallel processing. Therefore, Google Colab was utilized in this study. Google Colab is a cloud-based Jupyter notebook platform that provides integrated GPU and TPU acceleration. The environment runs on Ubuntu and provides access to NVIDIA Tesla K80 GPUs with 12 GB of GPU memory. Additionally, it includes a Python runtime along with pre-installed libraries and packages necessary for executing deep learning applications.
\subsubsection{Hyperparameter Tuning}
\begin{table}[h]
\centering
\caption{Hyperparameter Settings}
\begin{tabular}{|l|c|}
\hline
\textbf{Hyperparameter} & \textbf{Value} \\ \hline
Train/Validation/Test Split Ratio & 0.2 \\ \hline
Random Seed & 42 \\ \hline
Batch Size & 24 \\ \hline
Learning Rate & 2e-5 \\ \hline
Number of Epochs & 40 \\ \hline
\end{tabular}
\label{tab:hyperparameters}
\end{table}

Table \ref{tab:hyperparameters} presents the hyperparameter settings used for training the proposed deep learning model. The dataset was split into training and test sets at a 0.2 ratio. Here, the same data is used for validation and testing. , with a random seed value of 42, were applied to ensure reproducibility of the experimental results. The model was trained with a batch size of 24, meaning 24 samples were processed per training iteration. A learning rate of 2e-5 was selected to control the step size during optimization. Furthermore, the model was trained for 40 epochs, allowing the network to iteratively learn patterns from the dataset and improve performance over multiple training cycles. An Early Stopping criterion was applied to prevent overfitting during training. The training process monitored the validation loss and stopped automatically if no improvement was observed for two consecutive epochs, while aiming to minimize the validation loss. The experimental results are intended to be shown in a large number of observations. Therefore, instead of dividing the data into three parts, the same data have been used as a test and validation set.

\subsection{Proposed Framework}
This subsection is about the proposed model. Figure \ref{propmethod}  is the visualization of the workflow of the proposed model. The significant parts of the proposed methodology are discussed below:

\begin{figure*}
    \centering
\includegraphics[width=0.8\linewidth]{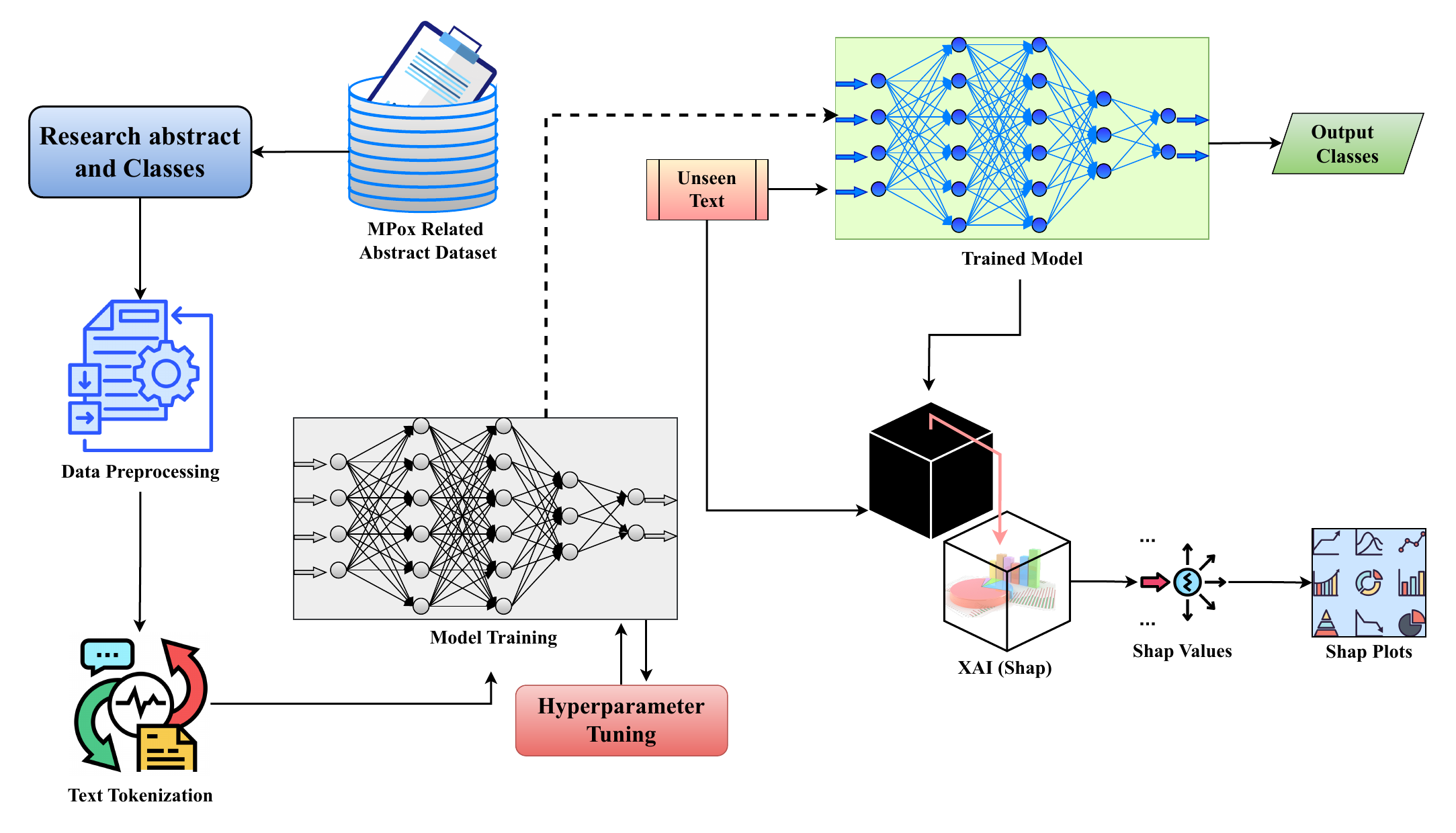}
    \caption{Workflow of the proposed Methodology}
    \label{propmethod}
\end{figure*}

\subsubsection{Data Collection and Preprocessing}

The research data collected from PubMed contains Biomedical and life science-related research articles. A total of 14590 articles were analyzed, and the final dataset was created with four distinct features named PMid, Title, Abstract, and labels. To extract data from PubMed, 9 keywords were used as search queries, and they are 'monkeypox', 'monkeypoxvirus', 'monkeypoxoutbreak', 'monkeypoxawareness', 'endmonkeypox', 'monkeypoxvaccine', 'mpox', 'monkeypoxprevention', 'monkeypoxnews'. As the dataset requires a large number of observations to train the deep learning model, no timeframe has been considered. However, the duplicate observations are discarded by checking the titles. 

\begin{figure}
    \centering
    \includegraphics[width=0.6\linewidth]{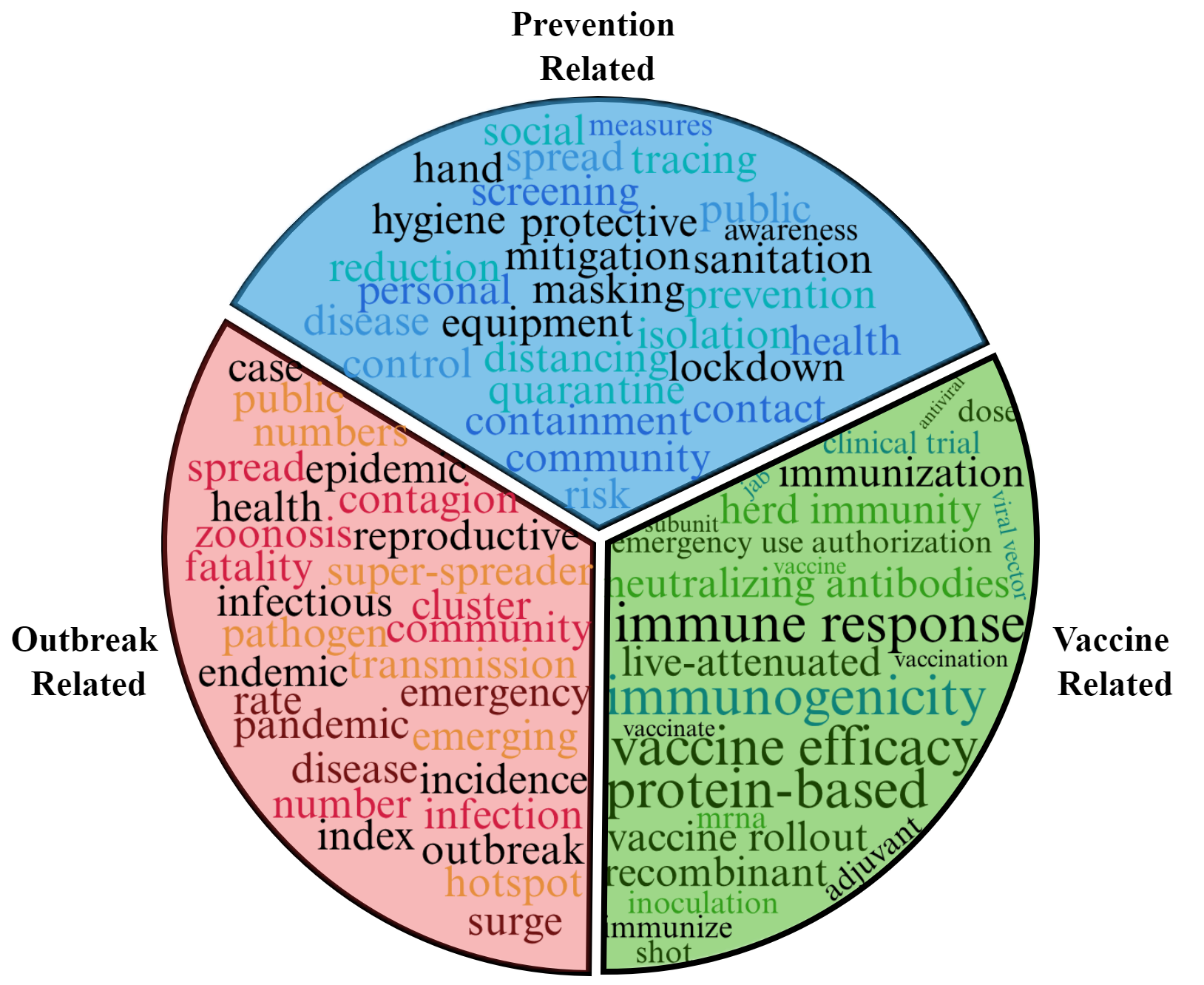}
    \caption{Word Cloud containing classwise keywords}
    \label{wc}
\end{figure}

\begin{figure}
    \centering
    \includegraphics[width=0.8\linewidth]{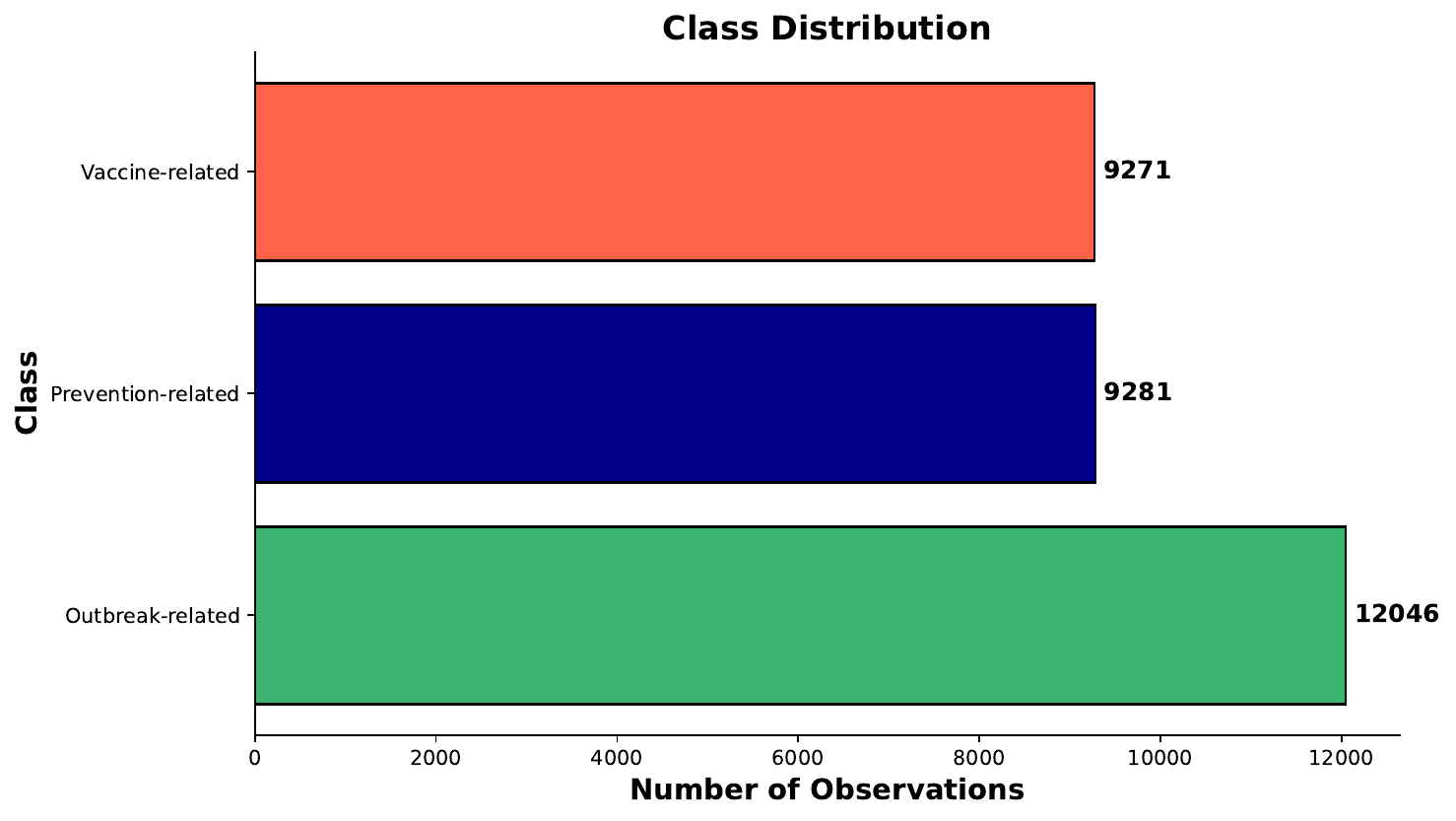}
    \caption{Number of observations for each classes}
    \label{classdist}
\end{figure}

The classes were selected based on the most frequently occurring keywords extracted from the abstracts. Initially, the number of classes was increased to four by introducing an additional treatment-related class. However, after conducting a manual analysis of the dataset, it was observed that the inclusion of the additional class introduced significant class imbalance, which negatively affected the performance of the classification models. Consequently, this research was ultimately conducted using three classes to ensure a more balanced dataset and improved model performance. The number of observations in each class is sketched in Figure \ref{classdist}. The outbreak-related class has the highest number of 12046 observations. The other two classes, Prevention-related and Vaccine-related classes, have 9281 and 9271 observations.

After collecting the corpus, the text and word frequency are analyzed. Based on the existing text, keywords are extracted, and three different classes are created vaccine: vaccine-related, outbreak-related, and prevention-related. The author manually executes this entire topic modeling process. Finally, based on the classwise keywords, the research articles are classified, and it is found that multiple papers can be categorized with more than one class. Figure \ref{wc} shows the word cloud, including the keywords for different classes.

\begin{figure}
    \centering
    \includegraphics[width=0.5\linewidth]{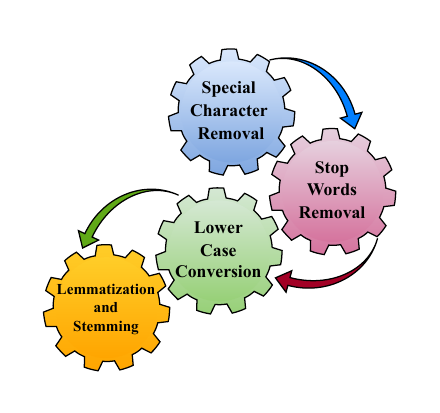}
    \caption{The preprocessing steps followed in the paper }
    \label{data prep}
\end{figure}

Data preprocessing techniques contribute to the model's performance. Therefore, different data preprocessing techniques have been applied to our text data. These techniques include removing special characters and stop words, changing upper-case characters to lower-case, and finally, lemmatization and stemming. Figure \ref{data prep} depicts the data preprocessing techniques used in this paper.

\subsubsection{Model Training and Evaluation}
After preprocessing the text, it has been tokenized using the appropriate transformer's tokenizer. The labels have been encoded into binary arrays, where a value of 1 indicates that a label is valid for the corresponding text, while a value of 0 indicates it is not.
Next, the tokenized text and encoded labels are utilized to train the transformer model. The performance of these models is compared using various evaluation metrics, including Binary Accuracy (Equation \ref{bin acc}), Micro F1 Score (Equation \ref{eq:maf2}), and Macro F1 score (Equation \ref{eq:maf1}). These metrics are used to conduct the performance comparison.

\begin{equation}
\label{bin acc}
    Accuracy=\frac{TN+TP}{TN+TP+FN+FP}
\end{equation}

\begin{equation}
    \label{eq:maf1}
Macro\ Average\ F1\ Score=\frac{2\times P_{MA}\times R_{MA}}{P_{MA}+ R_{MA}}
\end{equation}

\begin{equation}
    \label{eq:maf2}
Micro\ Average\ F1\ Score=\frac{\overline{TP}}{\overline{TP}+\frac{1}{2}(\overline{FP}+\overline{FN})}
\end{equation}

$\overline{TP}$ (net True Positive), $\overline{TN}$ (net True Negative), $\overline{FP}$ (net False Positive), and $\overline{FN}$ (net False Negative) are found from the confusion matrix of each class. Then, the obtained values are averaged to calculate the net values.
$P_{MA}$ and $R_{MA}$ stands for the macro average precision and recall. These values are obtained by averaging all the classes' precision and recall values.
\subsubsection{Model Explainability}
For the model's explainability, the SHAP algorithm is used. SHAP has shown which word features are essential in making the prediction and which words are acting negatively for the prediction. 
\section{Results and Discussion}
\subsection{Results}
\begin{figure}
    \centering
    \includegraphics[width=0.7\linewidth]{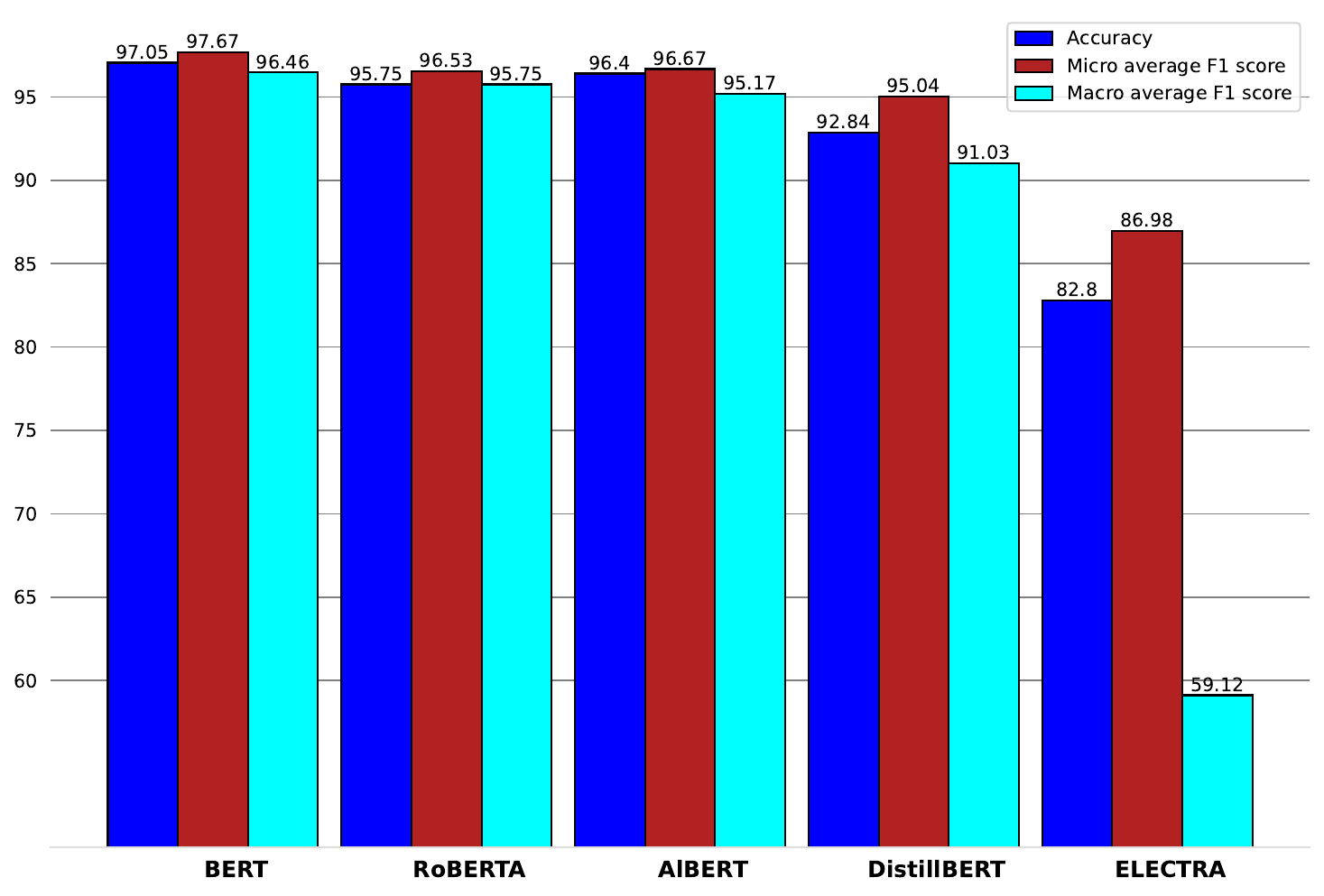}
    \caption{Comparing the Evaluation Metrics(Accuracy, Micro Average F1 Score, Macro Average F1 Score) for different transformer models}
    \label{comp}
\end{figure}

Figure \ref{comp} indicates a bar graph that compares the performance of different transformer-based models. The graph shows how well five different AI models (BERT, RoBERTa, ALBERT, DistilBERT, and ELECTRA) perform in extracting relationships between entities (like names, dates, etc.) from text. The three bars for each model represent Accuracy, Micro Average F1 Score, and Macro Average F1 Score. In simple terms, BERT and RoBERTa are the top performers, achieving the highest scores across all three metrics. Specifically, BERT has an Accuracy of 97.05\%, a Micro Average F1 Score of 97.67, and a Macro Average F1 Score of 96.46. RoBERTa scores 95.75, 96.53, and 95.75 in the same categories, respectively. On the other hand, ELECTRA doesn't do as well, especially in Macro Average F1 Score, with scores of 82.8, 86.98, and 59.12. This helps in understanding which models are better at processing and understanding text in this context.

Table \ref{tab:overall_multilabel_metrics} presents the overall multilabel classification performance of the proposed BERT model. The model achieved a Hamming Accuracy of 0.9332 and a low Hamming Loss of 0.0668, indicating that most labels were predicted correctly with minimal classification errors. Furthermore, the model attained high values for Micro Precision (0.9599), Micro Recall (0.9441), and Micro F1-score (0.9519), demonstrating strong overall predictive capability across all labels. Similarly, the Macro Precision, Recall, and F1-score values of 0.9597, 0.9411, and 0.9498, respectively, indicate that the model performs consistently across classes without significant bias toward any particular category. These findings confirm the effectiveness and robustness of the proposed BERT-based framework for multilabel classification of Mpox research articles.

\begin{table}[htbp]
\centering
\caption{Overall Multilabel Classification Performance}
\label{tab:overall_multilabel_metrics}
\begin{tabular}{lc}
\hline
\textbf{Metric} & \textbf{Value} \\
\hline
Hamming Accuracy & 0.9332 \\
Hamming Loss & 0.0668 \\
Micro Precision & 0.9599 \\
Micro Recall & 0.9441 \\
Micro F1-score & 0.9767 \\
Macro Precision & 0.9597 \\
Macro Recall & 0.9411 \\
Macro F1-score & 0.9646  \\
\hline
\end{tabular}
\end{table}

Table \ref{tab:classwise_performance} presents class-wise evaluation results for Precision, Recall, F1-score, and Support across the three research categories. The Outbreak-related class achieved the highest performance, with a Precision of 0.9671, Recall of 0.9752, and F1-score of 0.9711, indicating that the model effectively identifies outbreak-related studies. The Prevention-related category also demonstrated strong performance, achieving an F1-score of 0.9336, although its recall value (0.8933) suggests that a small number of relevant instances were missed. Similarly, the Vaccine-related class achieved a Precision of 0.9343, a recall of 0.9550, and an F1-score of 0.9445, confirming the model's ability to accurately recognize vaccine-focused articles. Overall, the high class-wise performance metrics indicate that the proposed model can reliably classify Mpox research articles across all categories.

\begin{figure}
    \centering
    \includegraphics[width=1\linewidth]{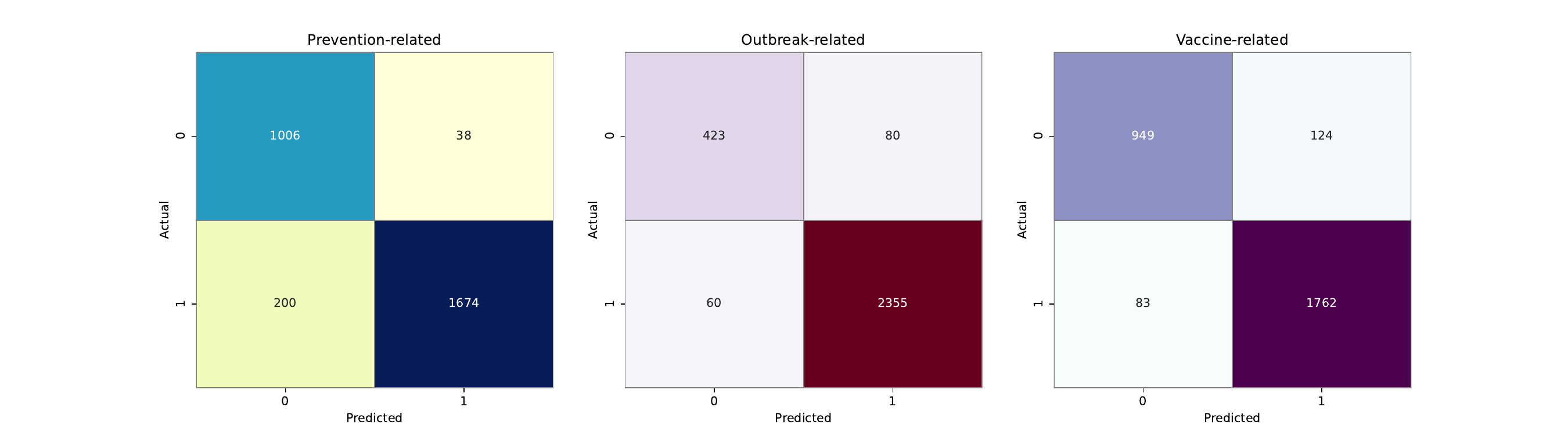}
    \caption{Classwise Confusion Matrix}
    \label{fig:classwise_cm}
\end{figure}

\begin{table}[htbp]
\centering
\caption{Class-wise Precision, Recall, F1-score, and Support}
\label{tab:classwise_performance}
\begin{tabular}{lcccc}
\hline
\textbf{Label} & \textbf{Precision} & \textbf{Recall} & \textbf{F1-score} & \textbf{Support} \\
\hline
Prevention-related & 0.9778 & 0.8933 & 0.9336 & 1874 \\
Outbreak-related & 0.9671 & 0.9752 & 0.9711 & 2415 \\
Vaccine-related & 0.9343 & 0.9550 & 0.9445 & 1845 \\
\hline
\end{tabular}
\end{table}
Figure \ref{fig:classwise_cm} illustrates the class-wise confusion matrices for the Prevention-related, Outbreak-related, and Vaccine-related categories generated by the BERT model. The confusion matrices provide a detailed view of the model's prediction performance by comparing actual and predicted labels. For the Prevention-related class, the model correctly identified 1,674 positive instances while misclassifying only a small number of samples. Similarly, for the Outbreak-related class, 2,355 positive instances were correctly classified, demonstrating excellent predictive performance with very few false predictions. In the Vaccine-related category, the model successfully classified 1,762 positive instances, indicating strong discriminative capability. Overall, the confusion matrices reveal that the BERT model achieves a high number of true positive predictions while maintaining relatively low false positive and false negative rates across all classes, further validating its effectiveness for multilabel classification.

\begin{figure}
    \centering    \includegraphics[width=0.7\linewidth]{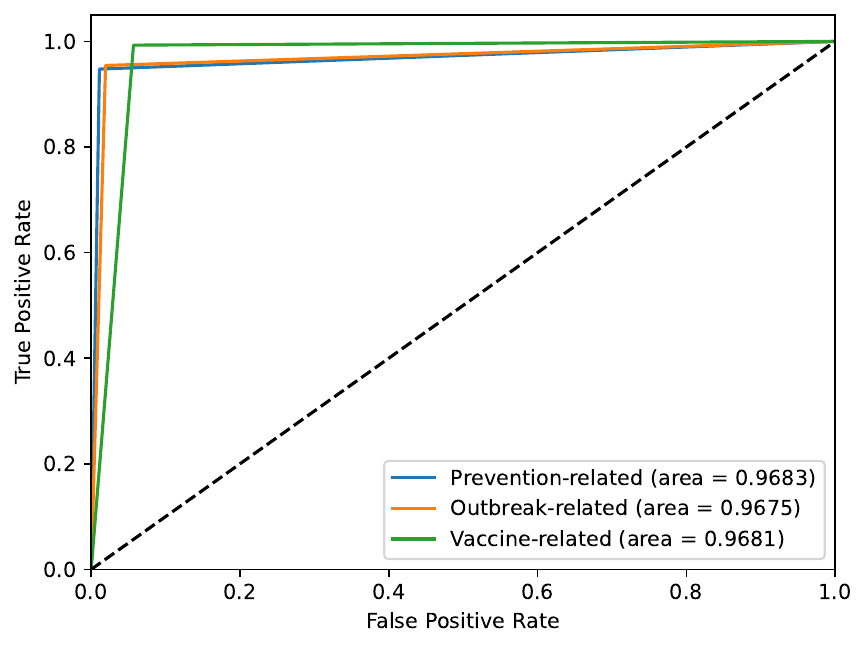}
    \caption{Receiver Operating Characteristic (ROC) Curve for Multilabel Classification with BERT}
    \label{MPox ROC}
\end{figure}

The ROC (Receiver Operating Characteristic) curve in figure \ref{MPox ROC} evaluated the performance of a multi-label classification BERT model. The ROC curve indicates the trade-off between the True Positive Rate and the False Positive Rate for each label in the classification task.

\begin{figure}
    \centering \includegraphics[width=1\linewidth]{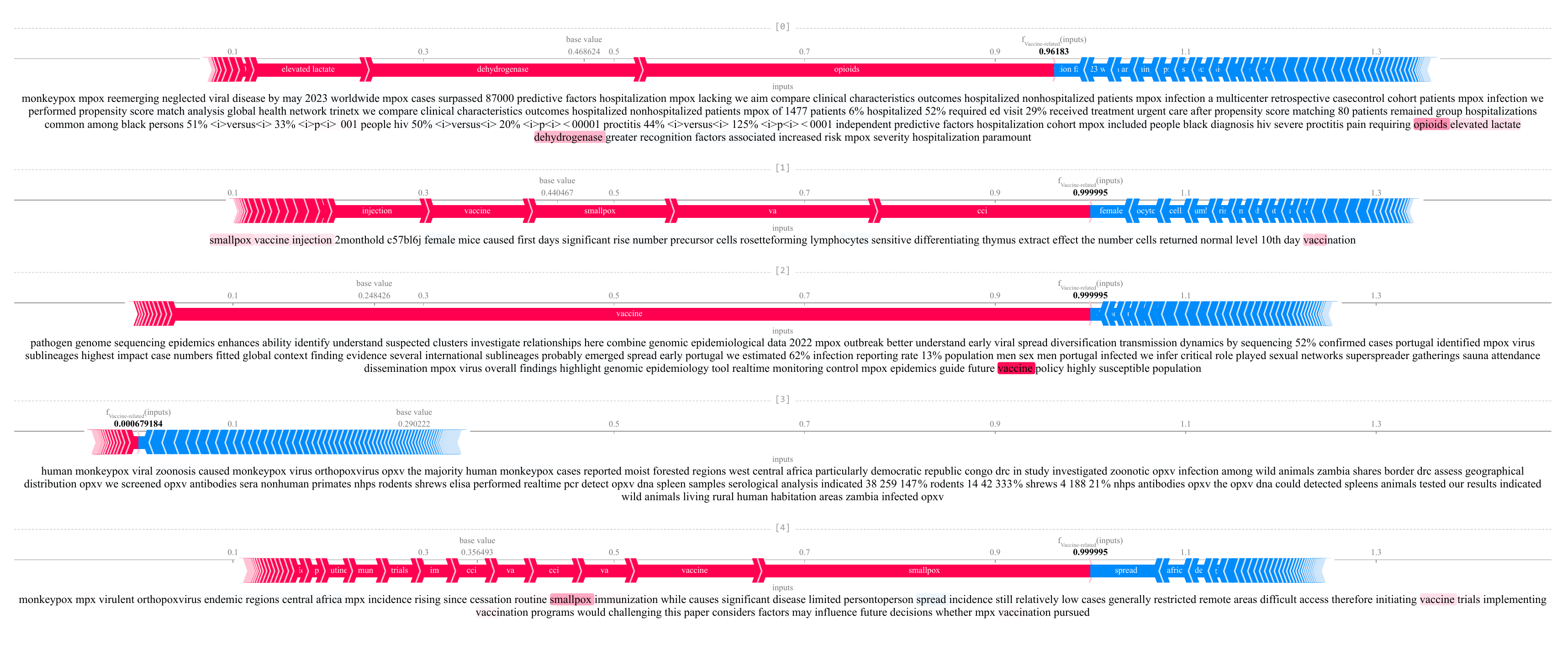}
    \caption{Word feature's contribution for class "Vaccine Related"}
    \label{outrel}
\end{figure}

\begin{figure}
    \centering
    \includegraphics[width=\linewidth, height=8cm]{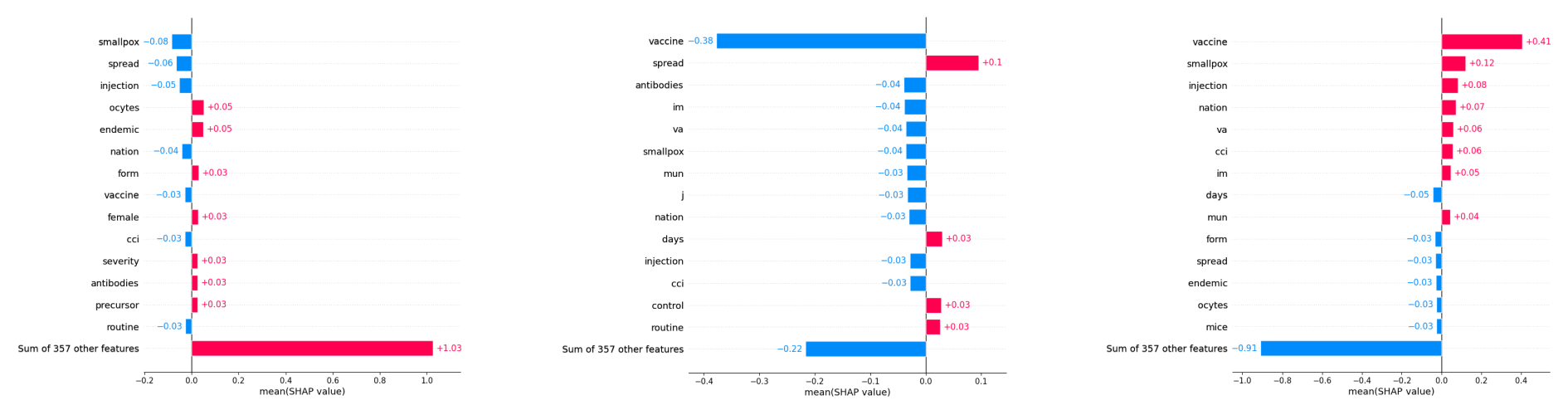}
    \caption{SHAP bar plots that contain important features for different classes.}
    \label{shap bars}
\end{figure}

In this graph, three ROC curves represent different labels:
\begin{itemize}
    
\item Prevention-related label: The blue line with an area under the curve (AUC) of 0.9683.
\item Outbreak-related label: The orange line with an AUC of 0.9675.
\item Vaccine-related label: The green line with an AUC of 0.9681.
\end{itemize}
AUC values close to 1 indicate excellent performance, meaning the model is very good at distinguishing between the different labels. The diagonal line is a random classifier with an AUC of 0.5, which is used as a baseline for comparison. Since the AUC values for all three labels are significantly higher than 0.5, the BERT model performs well in this multi-label classification task.
In summary, the high AUC values suggest that the BERT model is effective at correctly identifying and classifying each label in the dataset, making it a reliable choice for this type of task.
For model interpretability and explainability, SHAP is applied to the trained model. As it is a multilabel classification, SHAP text plots are generated for each class. Here, for the class "vaccine-related", the text plot has been added in Figure \ref{outrel}.
This SHAP plot illustrates how specific words and phrases influence the classification of research articles into the vaccine-related category. SHAP values quantify the contribution of each word to the model's final prediction, where red-highlighted words increase the likelihood of classification into the vaccine-related category, while blue-highlighted words decrease it. The thickness of the arrows represents how strongly each word influences the model's prediction. The base value represents the model's initial probability before considering specific words, while the final value $f_x$ (where xx refers to the article's category) shows the adjusted probability after key terms are taken into account.
The analysis highlights that words like "vaccine," "smallpox," "injection," "vaccination," and "trials" play a significant role in classifying research articles under the vaccine category. In many cases, the model starts with a probability between 0.2 and 0.4, but the presence of these terms significantly increases confidence, often pushing the final likelihood to 0.999995, indicating an almost certain classification. This suggests that research papers discussing vaccination strategies, smallpox vaccines, and immunization trials are strongly associated with the vaccine category.
For instance, in one example, the base value is 0.440467, but words like "vaccine" and "smallpox" remarkably increase the probability to 0.999995, confirming the model's confidence in classifying the paper as vaccine-related. Similarly, another example with a base value of 0.356493 sees a substantial increase in probability due to the presence of terms such as "trials," "vaccine," and "spread." These results demonstrate that articles related to vaccines, including the covering of vaccine trials, immunization efforts, and previous pox-related research, are key contributors to the vaccine category.
Notably, some words have little to no impact on vaccine classification. In one sample, zoonotic transmission-related terms (e.g., "wild animals," "OpVX," "primates") appear in the text but result in a very low $f_x$ value (0.000679184). This suggests that articles focusing on viral spread in animals or environmental reservoirs are less likely to be classified as vaccine-related as they do not directly address vaccination efforts.
Overall, the SHAP analysis confirms that the strongest predictors for the vaccine category include direct mentions of vaccines, immunization, smallpox, and clinical trials. The model effectively differentiates vaccine-related articles from those focusing on other aspects of Mpox, such as outbreaks or zoonotic transmission. These results demonstrate the model's effectiveness in identifying research papers based on vaccine-related terms, ensuring precise and dependable classification in multilabel tasks. 
Figure \ref{shap bars} depicts three different SHAP bar plots where the leftmost bar plot indicates the word importance for the Outbreak-related class, the middle one is for the Prevention-related class, and the remaining rightmost bar is for the vaccine-related category.
These three SHAP bar plots illustrate the contribution of different words to the model's prediction for classifying research articles into different categories. Each plot shows the mean SHAP values for individual words, indicating whether they increase (red) or decrease (blue) the probability of an article being classified under a particular category. Detailed explanation is given below:
\begin{itemize}

\item \textbf{Outbreak-related:} The word "routine" has the highest positive SHAP value (+1.03), meaning it strongly increases the likelihood of an article being classified as Outbreak-related. Words like "smallpox," "spread," and "injection" have negative SHAP values (-0.08, -0.06, -0.05), meaning they slightly decrease the probability of Outbreak-related classification. Other terms, such as "ocytes," "endemic," and "antibodies," have minor positive contributions but are not as influential as "routine."
\item \textbf{Prevention-related:} The word "vaccine" has a strong negative SHAP value (-0.38), meaning its presence actually decreases the probability of Prevention-related classification in this specific case. Other words like "spread," "antibodies," "smallpox," and "nation" also show slight negative contributions. Only a few terms, such as "nation" (+0.03) and "control" (+0.03), slightly increase the classification probability. The sum of 357 other features contributes negatively (-0.22), meaning many background words collectively reduce the likelihood of classification.
\item \textbf{Vaccine-related:} The word "vaccine" has the most substantial positive impact (+0.41), confirming that its presence greatly increases the probability of vaccine classification. Other important contributors include "smallpox" (+0.12), "injection" (+0.08), and "nation" (+0.07), indicating that articles discussing these topics are likely to be vaccine-related. Some words, like "days" (-0.05), "spread" (-0.03), and "endemic" (-0.03), slightly decrease the classification probability. The sum of 357 other features contributes a strong negative impact (-0.91), suggesting that less relevant words collectively reduce the probability of an article being classified in this category.
\end{itemize}

\begin{table}[h]
\centering
\caption{Average SHAP Importance per Class (Top 20 Tokens)}
\label{tab:shap_importance}
\begin{tabular}{lcccc}
\hline
\textbf{Token} & \textbf{Prevention-related} & \textbf{Outbreak-related} & \textbf{Vaccine-related} & \textbf{Overall Mean} \\
\hline
control       & 0.523 & 0.205 & 0.169 & 0.299 \\
spread        & 0.387 & 0.072 & 0.017 & 0.159 \\
vaccine       & 0.020 & 0.020 & 0.312 & 0.117 \\
endemic       & 0.033 & 0.085 & 0.061 & 0.060 \\
incidence     & 0.010 & 0.136 & 0.030 & 0.059 \\
rising        & 0.017 & 0.073 & 0.050 & 0.047 \\
va            & 0.018 & 0.015 & 0.097 & 0.043 \\
cci           & 0.012 & 0.014 & 0.075 & 0.034 \\
injection     & 0.013 & 0.049 & 0.033 & 0.032 \\
smallpox      & 0.024 & 0.012 & 0.052 & 0.029 \\
antibodies    & 0.012 & 0.013 & 0.052 & 0.026 \\
distribution  & 0.014 & 0.026 & 0.034 & 0.025 \\
mun           & 0.004 & 0.006 & 0.057 & 0.022 \\
im            & 0.004 & 0.006 & 0.055 & 0.022 \\
nosis         & 0.021 & 0.017 & 0.025 & 0.021 \\
nation        & 0.007 & 0.026 & 0.021 & 0.018 \\
early         & 0.042 & 0.004 & 0.007 & 0.018 \\
investigated  & 0.023 & 0.006 & 0.018 & 0.016 \\
effect        & 0.004 & 0.023 & 0.019 & 0.015 \\
transmission  & 0.016 & 0.018 & 0.010 & 0.015 \\
\hline
\end{tabular}
\end{table}

Table \ref{tab:shap_importance} displays the top 20 text tokens ranked by their average SHAP (SHapley Additive exPlanations) importance values across three distinct text classification categories: Prevention-related, Outbreak-related, and Vaccine-related, alongside their Overall Mean importance. Higher numerical values indicate that a specific token carries greater weight in driving the machine learning model's classification decisions for that particular class. For instance, the token "control" holds the highest overall importance ($0.299$), heavily influencing the Prevention-related category ($0.523$), while "vaccine" naturally serves as a dominant feature for the Vaccine-related class ($0.312$) but has negligible impact on the others. Ultimately, this feature attribution table illuminates which specific keywords the NLP model relies upon most heavily to differentiate between various public health contexts.

\subsection{Discussion}
This research has conducted unique but significant NLP research, where research articles are categorized. Moreover, the transparency and error analysis of the model was observed to understand how the classification was executed, from features to prediction. The practical implications of this work can be helpful for researchers. Before downloading or subscribing to a paper, researchers and academicians can find out that the article is related to the topic they are searching for. Moreover, automatically categorized papers will help researchers to collect papers and to save time during their study.

Here different transformer models are trained to find out the best model for this multilabel task.The evaluation metrics for multilabel classification of MPox research articles demonstrate clear performance differences among transformer models, including BERT, RoBERTa, ALBERT, DistilBERT, and ELECTRA. The three key metrics used for assessment are accuracy, micro-average F1 score, and macro-average F1 score.
Among all models, BERT achieves the highest performance, with an accuracy of 97.05\%, a micro-average F1 score of 97.67\%, and a macro-average F1 score of 96.46\%. These results indicate that BERT is highly effective in classifying MPox research articles across multiple labels. RoBERTa and ALBERT also exhibit strong performance, with only minor differences in their scores, suggesting they are competitive alternatives to BERT.
DistilBERT, a lightweight version of BERT, shows slightly lower performance, with an accuracy of 92.84\%, a micro-average F1 score of 95.04\%, and a macro-average F1 score of 91.03\%. Although it does not match BERT's performance, its efficiency and reduced computational requirements make it a viable option for classification tasks that prioritize faster inference times over absolute accuracy.
ELECTRA, however, performs significantly lower than the other models, with an accuracy of 86.98\%, a micro-average F1 score of 82.8\%, and a macro-average F1 score of 59.12\%. The sharp decline in the macro F1 score suggests that ELECTRA struggles to balance classification across all labels, likely due to difficulties in handling underrepresented classes.
Overall, these findings establish BERT as the most effective model for MPox research article classification, followed closely by RoBERTa and ALBERT. While DistilBERT offers a reasonable trade-off between performance and efficiency, ELECTRA appears less suitable for this specific task. Future research could explore fine-tuning these models further or incorporating domain-specific pretraining to enhance classification performance.

The graph with the ROC curve for the multilabel classification task further highlights the strong performance of BERT across three distinct categories: Prevention-related, Outbreak-related, and Vaccine-related research. The AUC (Area Under the Curve) values for all three classess are notably high, with Prevention-related at 0.9683, Outbreak-related at 0.9675, and Vaccine-related at 0.9681. These values confirm that BERT is highly effective in distinguishing between relevant and irrelevant instances for each category.
The ROC curves are positioned near the upper-left corner of the curve, showing a high true positive rate (sensitivity) and a low false positive rate. This suggests that the model effectively classifies relevant research articles while minimizing misclassification.
The Vaccine-related class (AUC = 0.9681) a bit outperforms the other two classes, indicating that BERT is particularly proficient at identifying vaccine-related research papers. However, the marginal differences in AUC values suggest that the model maintains a well-balanced classification capability across all categories. These results further validate the effectiveness of BERT in handling multilabel classification tasks for MPox research articles.

The SHAP (SHapley Additive exPlanations) text plot provides valuable insights about the BERT model in making decisions for classifying MPox research articles. The visualization highlights different words that strongly influence the model's predictions, with red denoting words that increase the likelihood of classification and blue representing words that decrease it.
A key observation is that specific terms such as "vaccine," "smallpox," "injection," "opioids," and "elevated lactate" significantly impact classification. For instance, "vaccine" and "smallpox" act as strong positive indicators of the vaccine-related class, while "opioids" and "elevated lactate" may be associated with the outbreak-related class, possibly due to their link to severe cases requiring hospitalization. The presence of these terms in red suggests that the model effectively associates them with relevant MPox-related topics.
The SHAP plot also reveals the context-sensitive nature of BERT, as some words can either increase or decrease classification probability depending on their surrounding context. For example, while "vaccine" is positively contributing to classification in vaccine related category, it is decreasing classification probability in prevention related class, suggesting that the model considers the broader textual context rather than relying solely on individual keywords.
Another notable observation is that the sum of 357 other features often contributes negatively to classification probability for prevention related and vaccine related class. However these 357 features provide positive influence for thr outbreak related class. This observation emphasizes that not only individual words are important, but also the structure and contents of a research article also impacting the predictions. 
The model does not rely solely on single term but evaluates multiple contextual elements to make accuracte decision.
The interpretability and explainability provided by SHAP is significant for understanding the AI-driven classification models, particularly in scientific research. By visualizing essential features why the model assigns specific labels to research abstracts, domain experts can validate predictions, identify biasness of the model, and redefine the model as needed. This transparency enhances the credibility of AI-driven classification and ensures that important research findings are categorized accurately.

Looking back at the research questions that guided this work, the results answer most 
of them fairly clearly. BERT came out on top for multi-label MPox classification, 
beating RoBERTa, ALBERT, DistilBERT, and ELECTRA across every metric, with 97.05\% 
accuracy, a 97.67\% micro-F1 score, and a 96.46\% macro-F1 score, while RoBERTa and 
ALBERT trailed only slightly, and ELECTRA fell apart noticeably on the underrepresented 
classes. The SHAP analysis, meanwhile, showed that the model isn't just 
keyword-spotting: terms like ‘vaccine’, ‘smallpox’, ‘injection’, and ‘trials’ 
consistently pushed predictions toward the vaccine-related class, ‘routine’ turned 
out to be the strongest driver for outbreak-related predictions, and ‘control’ and 
‘spread’ did similar work for prevention-related articles, but the same word could 
push a prediction in opposite directions depending on context, which points to real 
contextual reasoning rather than simple pattern matching. The challenge of 
overlapping topics was handled on two fronts, using a genuinely multi-label setup with 
sigmoid outputs instead of forcing single-label predictions, and trimming the label 
set from four categories down to three after an added treatment-related class 
introduced enough imbalance to hurt performance, which suggests overlapping labels 
were as much a data-curation problem as a modeling one. And on the practical side, 
pairing strong classification accuracy with SHAP-based explanations seems genuinely 
useful for public health work, since researchers and policymakers could use a tool 
like this to sort through MPox literature faster, while the visibility into the 
model's reasoning gives domain experts something to check the predictions against 
rather than having to trust a black box outright.

\section{Conclusion}
This study demonstrates the successive performance of transformer-based models, particularly BERT, in the multilabel classification of MPox research articles. BERT achieved the highest accuracy and F1 scores, confirming its suitability for this task. RoBERTa and ALBERT also performed well, while DistilBERT provided a viable trade-off between efficiency and accuracy. However, ELECTRA showed significantly lower performance than other models.
The ROC curve analysis further highlights BERT's strong predictive ability, with high AUC scores across all three classification categories—Prevention, Outbreak, and Vaccine. This suggests that BERT is highly reliable in distinguishing between different research topics related to MPox. SHAP text and bar plots provided transparency about the model's decision-making process by identifying words that contribute to the prediction and focusing on the importance of context in multilabel classifications. Overall, this research reinforces the potential of transformer-based models in organizing and analyzing scientific literature. Future work could focus on further fine-tuning, domain-specific pretraining, or integrating additional linguistic and contextual features to enhance classification accuracy. By leveraging advanced NLP techniques, researchers can streamline information retrieval in MPox-related studies, ultimately aiding in public health decision-making and research advancements.

\section*{Funding}
This research was conducted without any direct funding from public, commercial, or non-profit organizations.
\section*{Competing Interest and Ethics declarations}
\textbf{Competing interests:}
The authors declare no competing interests.

\textbf{Ethics approval and consent to participate:} Not applicable. This study does not involve human or animal subjects, and therefore, no ethical approval was required. All authors have reviewed and approved the submission.

\textbf{Declaration of Use Artificial Intelligence (AI):} No artificial intelligence (AI) tools or technologies were used in the preparation of this manuscript.

\end{document}